\documentclass[conference]{IEEEtran}
\IEEEoverridecommandlockouts
\usepackage{amsmath,amssymb,amsfonts}
\usepackage{algorithmic}
\usepackage[ruled]{algorithm2e}
\usepackage{array}
\usepackage{textcomp}
\usepackage{stfloats}
\usepackage{url}
\usepackage{color}
\usepackage{verbatim}
\usepackage{graphicx}
\usepackage{tabularx}
\usepackage{multirow}
\usepackage{array} 
\usepackage{booktabs}
\usepackage{graphicx}
\usepackage{parskip}

\newtheorem{definition}{Definition}
\usepackage[numbers]{natbib}
\usepackage{bm}
\usepackage{mathtools}

\usepackage{subfigure}
\usepackage{enumitem}

\usepackage{bibspacing}
\graphicspath{ {./fig/} }

\def\BibTeX{{\rm B\kern-.05em{\sc i\kern-.025em b}\kern-.08em
    T\kern-.1667em\lower.7ex\hbox{E}\kern-.125emX}}
\begin{document}

\title{Rapid and Continuous Trust Evaluation for Effective Task Collaboration Through Siamese Model\\
}

\author{\IEEEauthorblockN{Botao Zhu and Xianbin Wang}
\IEEEauthorblockA{Dept. of Electrical and Computer Engineering, Western University,
London, Ontario N6A 3K7 CANADA \\}   
}

\maketitle

\begin{abstract}




Trust is emerging as an effective tool to ensure the successful completion of collaborative tasks within collaborative systems. However, rapidly and continuously evaluating the trustworthiness of collaborators during task execution is a significant challenge due to distributed devices, complex operational environments, and dynamically changing resources. To tackle this challenge, this paper proposes a Siamese-enabled rapid and continuous trust evaluation framework (SRCTE) to facilitate effective task collaboration. First, the communication and computing resource attributes of the collaborator in a trusted state, along with historical collaboration data, are collected and represented using an attributed control flow graph (ACFG) that captures trust-related semantic information and serves as a reference for comparison with data collected during task execution. At each time slot of task execution, the collaborator's communication and computing resource attributes, as well as task completion effectiveness, are collected in real time and represented with an ACFG to convey their trust-related semantic information. A Siamese model, consisting of two shared-parameter Structure2vec networks, is then employed to learn the deep semantics of each pair of ACFGs and generate their embeddings. Finally, the similarity between the embeddings of each pair of ACFGs is calculated to determine the collaborator's trust value at each time slot. A real system is built using two Dell EMC 5200 servers and a Google Pixel 8 to test the effectiveness of the proposed SRCTE framework. Experimental results demonstrate that SRCTE converges rapidly with only a small amount of data and achieves a high anomaly trust detection rate compared to the baseline algorithm.

\end{abstract}


\section{Introduction}
Effective collaboration is becoming crucial yet more challenging in diverse systems due to the increasing system scale, resource constraints, and interdependence among distributed facilities involved. Trust evaluation among distributed devices has become an essential mechanism to ensure effective task completion~\cite{b1}. To achieve trust-based collaboration, a task initiator needs to evaluate the trust levels of potential collaborators based on its expectations regarding their capabilities to complete the initiator's tasks before assigning collaborative tasks~\cite{b2}. These capabilities include factors essential to task completion, such as computing resources, communication resources, and historical performance~\cite{bzhu}. 


However, collaborators selected as trustworthy before the task begins may not maintain their trustworthiness throughout the task execution. Since the resources and behaviours of collaborators are dynamic, their trustworthiness may fluctuate over time. For example, if the quality of communication links between the task initiator and collaborators deteriorates, indicating unreliable communication resources, the task transmission process may experience increased delays, jitter, and packet loss~\cite{b3}. Such poor communication links reduce collaborators' capabilities below those assessed in a trusted state~\cite{qhan}, leading to a decline in their trust. Similarly, if a collaborator engages in other tasks while executing the task from the task initiator, multiple tasks may compete for computing resources such as CPU, memory, and storage~\cite{b4}. When resources are constrained, the collaborator's response time may increase, potentially leading to the failure of resource-intensive tasks~\cite{b5}. Under such conditions, the collaborator's computing performance is likely to be lower than its performance in a trusted state, resulting in a diminished trust level. Furthermore, the task completion effectiveness of collaborators plays a crucial role in determining their trustworthiness. If the task completion effectiveness of collaborators falls short of the task initiator's expectations, the initiator's trust in them decreases. Based on the above analysis, it is important for collaborators to maintain continuous trust for task completion. Therefore, it is essential to evaluate the trustworthiness of collaborators swiftly and continuously during task execution in collaborative systems.

Due to the complex effects of changing resources and task execution outcomes, rapidly and continuously assessing collaborators' trustworthiness during dynamic collaboration poses a significant challenge. Thus, an urgent need arises for a new approach to evaluate collaborators' trustworthiness during task execution to ensure effective task completion. To address this challenge, this paper proposes a Siamese-enabled rapid and continuous trust evaluation framework (SRCTE). The proposed SRCTE framework begins by collecting the collaborator's resource attributes and historical collaboration data in a trusted state, and uses an attributed control flow graph (ACFG) to represent their semantic information. This ACFG serves as a reference for comparison with data collected during task execution. Next, the framework gathers the collaborator's resource attributes and task collaboration effectiveness at each time slot of task execution and represents their semantic information using an ACFG. Furthermore, a straightforward Siamese-enabled model is employed to compute the similarity between each pair of ACFGs. A high similarity value indicates that the collaborator remains trustworthy, while a low value suggests a lack of trustworthiness. If the collaborator is deemed untrustworthy at any time slot, the task initiator can promptly terminate the collaboration. The contributions of this paper are summarized as follows:
\begin{itemize}[leftmargin=*]
    \item 
    This paper proposes using trust as a mechanism to ensure effective task execution in collaborative systems, creatively transforming trust assessment into the dynamic, real-time evaluation of resources and behaviours of collaborators. Additionally, it presents a straightforward yet innovative SRCTE framework that enables rapid and continuous trust evaluation of collaborators in dynamic operational environments.

\item
    To achieve rapid trust evaluation, the proposed SRCTE leverages the collaborator's communication and computing resource attributes in a trusted state, as well as historical collaboration data, as a reference for comparison, and employs an ACFG to capture the trust-related semantic information in these data. It then continuously collects the collaborator's communication and computing resource attributes at each time slot of task execution, along with task completion effectiveness, and uses an ACFG to represent the trust-related semantic information for each time slot.

\item 
   A simple Siamese model, consisting of two shared-parameter Structure2vec networks, is employed to learn the deep semantics of each ACFG pair composed of the ACFG in the trusted state and the ACFG at each time slot. The similarity between each pair of ACFGs is then calculated to assess the trust value of the collaborator based on the learned embeddings of the ACFGs.

\item 
   The proposed framework is easily deployed in a real collaborative system to evaluate its effectiveness. Experimental results demonstrate that it converges rapidly with a small amount of training data and achieves a high anomaly trust detection rate compared to the baseline algorithm.


\end{itemize}

\section{System Model and Problem Formulation}
This paper considers a collaborative computing system consisting of a group of devices, denoted as $\bm{A}=\{a_1,\dots, a_I\}$, and an edge server responsible for managing and coordinating their interactions. Each device can serve as a task initiator, creating collaborative tasks and identifying trusted devices to assist with task execution. A collaborative task is composed of multiple subtasks, represented as $\bm{B}=\{b_1,\dots,b_M\}$. The trust of collaborators in the collaborative computing system is defined as follows: 

\begin{definition}[Trust in the collaborative computing system] 
\textit{Assuming $a_i \in \bm{A}$ is the task initiator generating a task $\bm{B}$, it evaluates the likelihood of each potential collaborator $a_j \in \bm{A}$ completing $\bm{B}$ based on $a_j$'s communication and computing resource attributes, as well as its task completion effectiveness. This likelihood is treated as $a_i$'s trust in $a_j$, represented as:
\begin{align}
    T_{a_i, a_j} = Trust(\bm{R}^{\text{com}}, \bm{R}^{\text{cmp}},\bm{R}^{\text{efc}}),
\end{align}
where $\bm{R}^{\text{com}}$ is the set of communication resource attributes of $a_j$, $\bm{R}^{\text{cmp}}$ represents the set of computing resource attributes, and $\bm{R}^{\text{efc}}$ indicates its task completion effectiveness.}
\end{definition}

Assuming $a_j$ is in a trusted state at time $t_0$, its computing resource attributes $\bm{R}^{\text{cmp}} (t_0)$, communication resource attributes $\bm{R}^{\text{com}}(t_0)$, and historical task completion effectiveness $\bm{R}^{\text{efc}}(t_0)$ are collected. The trust value of $a_i$ towards $a_j$ is then calculated as $T_{a_i, a_j} (t_0) = Trust(\bm{R}^{\text{com}}(t_0), \bm{R}^{\text{cmp}}(t_0),\bm{R}^{\text{efc}}(t_0))$. Subsequently, $a_i$ sequentially transmits each subtask in 
$\bm{B}$ to $a_j$, which can receive the subtasks while simultaneously executing those that have already been received. The task execution process is assumed to be divided into $N$ equal time slots, denoted as $\{ t_1,\dots,t_n, \dots,t_N\}$.
During each time slot $t_n$, we gather $a_j$'s computing resource attributes $\bm{R}^{\text{cmp}}(t_n)$, communication resource attributes $\bm{R}^{\text{com}}(t_n)$, and task completion effectiveness $\bm{R}^{\text{efc}}(t_n)$ in real-time. These attributes may vary across time slots, resulting in corresponding changes in trust. If the transition of $a_j$ from trustworthy to untrustworthy is not detected in a timely manner, it could lead to task failure. Therefore, to ensure a trustworthy collaboration process for effective task completion, it is necessary to conduct a rapid and continuous trust evaluation for device $a_j$ throughout the entire task execution process. The definition of a trusted collaborator in the collaboration process is as follows 

\begin{definition} [Trusted collaborator in the collaboration process]
    \textit{
    If the ratio of the trust value $T_{a_{i},a_{j}} (t_n)$ of $a_j$ at time slot $t_n$ to its trust value $T_{a_{i},a_{j}} (t_0)$ in a trusted state is not less than the threshold $\Delta$, $a_j$ is considered trustworthy at time slot $t_n$, which is given by}
    \begin{align}
        \label{progresstrust}
        T_{a_{i},a_{j}} (t_n)/T_{a_{i},a_{j}} (t_0) \ge \Delta. 
    \end{align}
\end{definition}
Since trust calculation involves various factors, it is challenging to accurately determine the collaborator's trust value during task execution. However, if trust can be inferred from changes in the collaborator's resource attributes and task completion effectiveness, obtaining the trust value becomes straightforward. To achieve this goal, the SRCTE framework is presented in this paper to quickly and continuously assess the trustworthiness of the collaborator.


\section{Siamese-Enabled Rapid and Continuous Trust Evaluation Framework}



  The implementation of the proposed SRCTE framework mainly includes the following steps:
  \subsubsection{Trust-related data collection of the collaborator} SRCTE collects the resource attributes and historical collaboration information of the collaborator in a trusted state, then continuously gathers its resource attributes and task completion effectiveness at each time slot during task execution.
  
  \subsubsection{Trust-related semantic representation of the collected data}  ACFGs are used to represent the trust-related semantic information of the collected data. Specifically, the data in a trusted state and the data from each time slot are each represented using an ACFG, forming a pair of ACFGs.

  \subsubsection{Trust evaluation of the collaborator at each time slot by comparing the similarity of each pair of ACFGs} A Siamese model is employed to learn the embeddings of each pair of ACFGs, and a similarity calculation method is used to evaluate the trust value of the collaborator based on these embeddings.
  
\begin{figure}[!]
\centering
\includegraphics[scale=0.9]{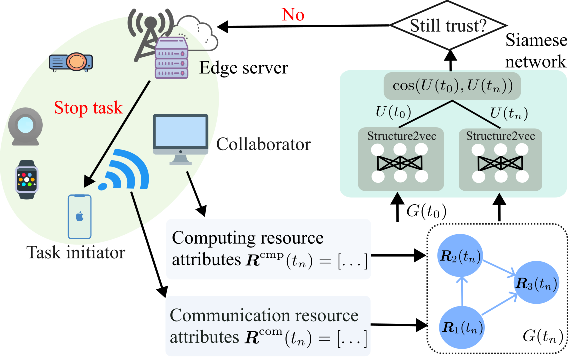}
\caption{The proposed SRCTE framework in the collaborative computing system.}
\label{framework}
\end{figure}

  SRCTE achieves rapid and continuous trust evaluation by simplifying the complex trust calculation into a comparison of the collaborator's resources and task completion effectiveness. It is illustrated in Fig.~\ref{framework} and presented in Algorithm~\ref{siamese_algorithm}, and its specific implementation details are described below.

\vspace{-0.04 in}

\begin{algorithm}[t!] 
	\caption{{The SRCTE framework}}
	\label{siamese_algorithm}
	\begin{algorithmic}[1]
		\renewcommand{\algorithmicrequire}{\textbf{Input:}}
		\REQUIRE Task initiator $a_i$, selected collaborator $a_j$
            \renewcommand{\algorithmicrequire}{\textbf{Output:}}
            \STATE collect  $\bm{R}^{\text{com}}(t_0), \bm{R}^{\text{cmp}}(t_0)$, and $\bm{R}^{\text{efc}}(t_0)$ of $a_j$ at the trusted state $t_0$ \\
            \STATE generate ACFG $G(t_0) = (\mathcal{A}(t_0), \mathcal{E}(t_0))$ based on the collected information \\
            \STATE assign task $\bm{B}$ to $a_j$ \\
            \FOR{each time slot $t_n$ =  $t_1 : t_N$}
                \STATE collect $\bm{R}^{\text{com}}(t_n), \bm{R}^{\text{cmp}}(t_n)$, and $\bm{R}^{\text{efc}}(t_n)$ of $a_j$ \\
                \STATE generate ACFG $G(t_n) = (\mathcal{A}(t_n), \mathcal{E}(t_n))$ \\
                \STATE input $G(t_0)$, $G(t_n)$ into two Structure2ve networks \\
                \FOR{each $G(t) \in \{G({t_0}), G({t_n})\}$}
                \FOR{$l = 1 : L$} 
                  \FOR{each $\bm{R}_k \in \mathcal{A}$}
                     \STATE  $u^{l}_{k} =  \tanh \left(W_1 \bm{R}_{k} + \sigma \left(\sum_{u_j \in \bm{g}_{\bm{R}_k}}u^{{l-1}}_{j}  \right) \right)$ \\
                  \ENDFOR
                \ENDFOR
                \STATE $U = W_2 \sum_{k \in \{1,2,3\}}  u^{L}_{k}$ \\
                \ENDFOR
                 \STATE $U(t_0), U(t_n)$ \\
                \IF{$\cos(U(t_0),U(t_n)) \ge \Delta$}
                    \STATE $a_j$ is trusted \\
                    \ELSE
                     \STATE $a_j$ is untrusted, and task $\bm{B}$ is terminated
                \ENDIF
            \ENDFOR
	\end{algorithmic}
\end{algorithm}

\subsection{Edge server-assisted just-in-time collaboration model}
To ensure efficient resource utilization, task initiator $a_i$ and collaborator $a_j$ adopt an edge server-assisted just-in-time collaboration model. Once $a_i$ selects $a_j$ as the collaborator, they register their collaboration, denoted as $[a_i, a_j]$, on the edge server. Upon completion or termination of the collaboration, they deregister to avoid unnecessary resource consumption. {Before the task begins, the edge server collects information $\{\bm{R}^{\text{com}}(t_0), \bm{R}^{\text{cmp}}(t_0), \bm{R}^{\text{efc}}(t_0)\}$ about $a_j$ in a trusted state.} During the task execution process, $a_j$ needs to continuously report its resource attributes and task completion effectiveness to the edge server. The information transmission between the devices and the edge server is assumed to be reliable.

The main functions of the edge server include: 1) collecting information $\{\bm{R}^{\text{com}}(t_n), \bm{R}^{\text{cmp}}(t_n), \bm{R}^{\text{efc}}(t_n)\}$ from $a_j$ at each time slot $t_n$; 2) quickly inferring the trustworthiness of $a_j$ based on  $\{\bm{R}^{\text{com}}(t_n), \bm{R}^{\text{cmp}}(t_n), \bm{R}^{\text{efc}}(t_n)\}$; and 3) promptly notifying $a_i$ to terminate the task if $a_j$ is determined to be untrustworthy.

\subsection{Siamese-enabled rapid and continuous trust evaluation}


To evaluate trust, $\{\bm{R}^{\text{com}}(t_0), \bm{R}^{\text{cmp}}(t_0), \bm{R}^{\text{efc}}(t_0)\}$
is used as a reference and compared with $\{\bm{R}^{\text{com}}(t_n), \bm{R}^{\text{cmp}}(t_n), \bm{R}^{\text{efc}}(t_n)\}$ at each time slot to determine the trustworthiness of the collaborator. To facilitate this comparison, ACFGs are used to represent their trust-related semantic information. Additionally, a Siamese model is employed to learn the embeddings of each ACFG pair, capturing deep semantics, and outputs the collaborator's trust value by performing the similarity calculation between these embeddings.

\subsubsection{Trust-related semantic representation of communication, computing, and task completion effectiveness through ACFG} The comparison between $\{\bm{R}^{\text{com}}(t_n),\bm{R}^{\text{cmp}}(t_n), \bm{R}^{\text{efc}}(t_n)\}$ and $\{\bm{R}^{\text{com}}(t_0), \bm{R}^{\text{cmp}}(t_0), \bm{R}^{\text{efc}}(t_0)\}$ is not an independent comparison of their corresponding elements, but rather a comparison of them as a whole. However, accurately describing the influence of each element on trust is challenging. This is because the communication resource attributes $\bm{R}^{\text{com}}(t_n)$ affect computing, and both communication and computing attributes jointly impact the final task effectiveness $\bm{R}^{\text{efc}}(t_n)$. {In this research, we use ACFGs~\cite{b6} to represent the interdependencies and trust-related semantic information between them.} As shown in Fig.~\ref{framework}, an ACFG representing communication, computing, and task completion effectiveness at time slot $t_n$ is denoted as $G(t_n) = (\mathcal{A}({t_n}), \mathcal{E}({t_n}))$, where 
$\mathcal{A}(t_n) = \{\bm{R}_1(t_n), \bm{R}_2(t_n), \bm{R}_3(t_n)\}$ represents the set of vertices, corresponding to the elements in $\{\bm{R}^{\text{com}}(t_n),\bm{R}^{\text{cmp}}(t_n), \bm{R}^{\text{efc}}(t_n)\}$, and $\mathcal{E}(t_n) = \{e_{\bm{R}_1(t_n) \to \bm{R}_2(t_n)}, e_{\bm{R}_2(t_n) \to \bm{R}_3(t_n)}, e_{\bm{R}_1(t_n) \to \bm{R}_3(t_n)} \}$ is the set of directed edges. Similarity, $\{\bm{R}^{\text{com}}(t_0), \bm{R}^{\text{cmp}}(t_0), \bm{R}^{\text{efc}}(t_0)\}$ is also transformed into an ACFG $G({t_0}) = (\mathcal{A}(t_0),\mathcal{E}(t_0))$. Ultimately, the trust evaluation of $a_j$ at time slot $t_n$ is reduced to a similarity assessment between $G({t_n})$ and $G({t_0})$. The greater the similarity between $G({t_n})$ and $G({t_0})$, the more trustworthy the collaborator is deemed to be at time slot $t_n$.

\subsubsection{Siamese-enabled ACFG similarity calculation}

Due to the complexity of traditional graph similarity calculation algorithms, this research employs the Siamese model for fast ACFG similarity assessment. A typical Siamese model consists of two identical neural networks with shared weights that independently process two inputs and compare their outputs \cite{b7}. This architecture is particularly effective for tasks requiring similarity or dissimilarity assessments and excels in scenarios with limited data, making it ideal for real-time comparison between entities \cite{b8}. 

\begin{figure}[!]
\centering
\includegraphics[scale=0.96]{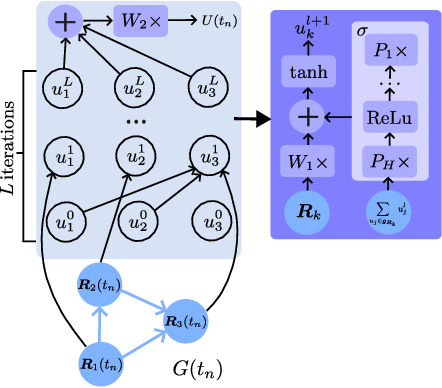}
\caption{Structure2vec generates the embedding for each ACFG.}
\label{structure2vec}
\end{figure}

Structure2vec~\cite{b9} is utilized to implement two identical neural networks within the Siamese model, as shown in Fig.~\ref{structure2vec}. It computes a $p$-dimensional feature $u$ for each vertex in the input graph and aggregates all vertex embeddings to generate the graph's embedding vector $U$. Initially, $G({t_0})$ and $G({t_n})$ are fed into Structure2vec, and the embedding vector $u^{0}_{k}, k=\{1,2,3\}$ at each vertex is randomly initialized. Both ACFGs then undergo $L$ iterations to generate the final graph embeddings. At each iteration $l$, vertex-specific embeddings at the last iteration are aggregated according to ACFG topology, and the new $u^{l}_{k}$ for each vertex in ACFG is generated taking into account both the vertex features and graph-specific characteristics. In particular, the vertex embedding $u_{k}$ is updated at the $(l+1)$-th iteration as follows:
\vspace{-0.05 in}
    \begin{align}
    u^{l+1}_{k} = F\left(\bm{R}_k, \sum_{u_j \in \bm{g}_{\bm{R}_k}} u^{l}_{j} \right), 
\end{align}
where $\bm{g}_{\bm{R}_k}$ is the set of neighbours of $\bm{R}_k$ in ACFG, and $F()$ is the nonlinear propagation function, which clearly defines a process in which vertex features $\bm{R}_k$ are propagated to other vertices. $F()$ is designed to have the following form
\begin{small}
    \begin{align} 
    F\left(\bm{R}_k, \sum_{u_j \in \bm{g}_{\bm{R}_k}} u^{l}_{j}  \right) = \tanh \left(W_1 \bm{R}_{k} + \sigma \left(\sum_{u_j \in \bm{g}_{\bm{R}_k}} u^{l}_{j}  \right) \right), 
\end{align}
\end{small}
where $W_1$ is a $d \times p$ matrix, $\tanh()$ represents the hyperbolic tangent function, and $\sigma()$ is the nonlinear transformation with an $H$ layers fully-connected neural network
\begin{align}
    \sigma(x) =  P_1 \times \text{ReLu}\left(P_2 \times \hdots \text{ReLu} (P_H x)\right),
\end{align}
where $P_h, h=1,\dots,H$, is a $p \times p$ matrix, $\text{ReLu}$ is the rectified linear unit, $\text{ReLu}(x) = \max\{0,x\}$. The final graph embedding of each ACFG is obtained by aggregating the vertex embeddings after $L$ iterations, which is given by
\begin{align}
    U = W_2 \sum_{k \in \{1,2,3\}}  u^{L}_{k},
\end{align}
where $W_2$ is a $p \times p$ matrix utilized to transform the graph embedding. Using the obtained embeddings $U(t_0)$ and $U(t_n)$, the similarity score between $G(t_0)$ and $G(t_n)$ is calculated by the cosine function
\begin{align}
   Simi(G(t_0),G(t_n)) &=  \cos(U(t_0), U(t_n)) \nonumber \\ 
   &= \frac{U(t_0) \cdot U(t_n)}{\parallel U(t_0)\parallel \parallel U(t_n)\parallel}.
\end{align}
This similarity score is viewed as the trust value of $a_j$ at time slot $t_n$, i.e., $T_{a_i,a_j}(t_n)$. {If $Simi(U(t_0),U(t_n)) \ge \Delta$, $a_j$ is trusted; otherwise, it is deemed untrusted. It is evident that the proposed SRCTE framework provides a straightforward approach for achieving rapid and continuous trust evaluation.}

\subsubsection{Learning parameters} 
To obtain the optimal result, the proposed SRCTE framework needs to be trained by a certain amount of data pairs of $\{\bm{R}^{\text{com}}(t_0), \bm{R}^{\text{cmp}}(t_0), \bm{R}^{\text{efc}}(t_0)\}$ and $\{\bm{R}^{\text{com}}(t_n), \bm{R}^{\text{cmp}}(t_n), \bm{R}^{\text{efc}}(t_n)\}$. Given $Q$ such data pairs, they are transformed into $Q$ ACFG pairs of $G_q(t_0)$ and $G_q(t_n)$. Each pair is associated with ground truth pairing information $y_q \in \{1, -1\}$, where $y_q = 1$ indicates that $G_q(t_0)$ and $G_q(t_n)$ are similar, otherwise dissimilar. Then, stochastic gradient descent is used to minimize the error function
\begin{align}
     \min_{W_1,P_1,\dots,P_H,W_2} \sum_{q=1}^{Q}\left(Simi \left(G_q(t_0), G_q(t_n)\right) - y_q \right)^2,
\end{align}
where $W_1,P_1,\dots,P_H,W_2$ are the trainable parameters. The generation of training data will be introduced in detail in the experimental section.


\section{Experiments}

\subsection{Experimental environment establishment}
\begin{figure}[!]
\centering
\includegraphics[scale=0.23]{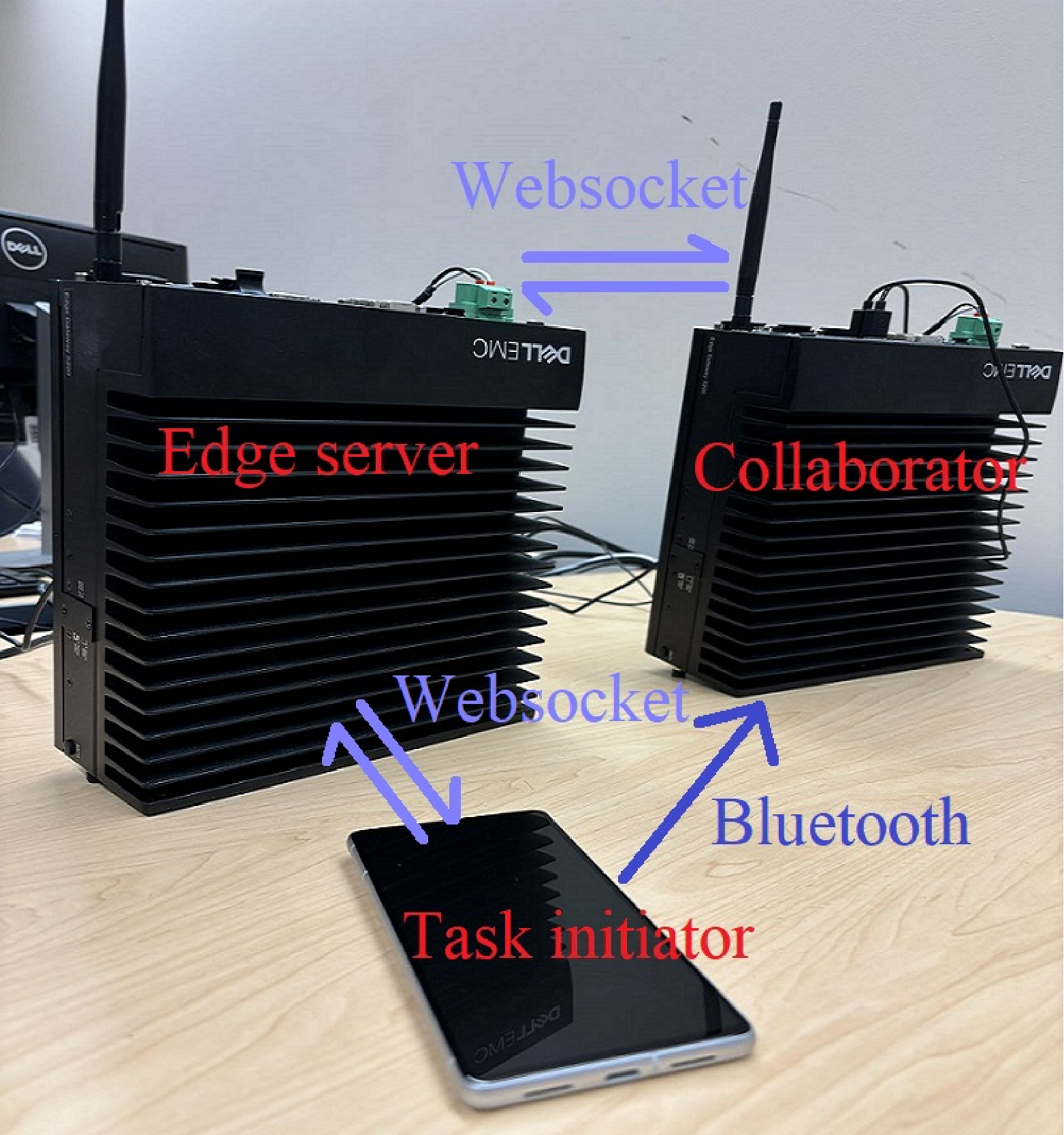}
\caption{A collaborative system with  two DELL EMC 5200
servers and a Google Pixel 8.}
\label{real_system}
\end{figure}

To validate the effectiveness of the proposed SRCTE framework, a collaborative computing system is established using two DELL EMC 5200 servers and a Google Pixel 8, as shown in Fig.~\ref{real_system}. One DELL EMC 5200 serves as the edge server, while the other simulates a collaborator. The Google Pixel 8 acts as the task initiator, generating a task comprising $M=200$ subtasks. Each subtask is a `.py' file containing a piece of Python code. The task initiator continuously sends subtasks to the task collaborator via Bluetooth, and the collaborator executes the code within each subtask upon receiving it. The task initiator, collaborator, and edge server are all connected to the Internet, and their communication and notification mechanisms are implemented using WebSocket. WebSocket allows for persistent connections, bidirectional communication, and efficient frame transmission, enabling real-time data pushing and receiving. 
The embedding size is set to $p = 64$. The number of iterations $L$ is set to $2$, and the number of embedding layers is set to $H=2$. The proposed SRCTE is implemented using Python and PyTorch.

\subsection{Dataset collection}
\label{datasetcollection}
All third-party applications are disabled, and the impact of system applications on communication and computing is disregarded. The total number of task executions is set to 2000, and Wireshark is used to capture data packets during task execution.

During each task execution, the collaborator's communication resource attributes $\bm{R}^{\text{com}}(t_n)$, computing resource attributes $\bm{R}^{\text{cmp}}(t_n)$, and task completion effectiveness $\bm{R}^{\text{efc}}(t_n)$ are calculated within each $t_n=0.5$ second time slot. The factors included in $\bm{R}^{\text{com}}(t_n)$ are: 1) the average delay of all packets received during the current time slot, $r^{\text{delay}}(t_n)$, and 2) the total number of subtasks received by the collaborator within that time slot, $r^{\text{total}}(t_n)$. For simplicity, we assume that the number of received subtasks is determined solely by communication. To capture $\bm{R}^{\text{cmp}}(t_n)$, we continuously collect the following information during each time slot: 1) the collaborator's overall CPU utilization $r^{\text{CPU}}(t_n)$, and 2) the average completion time of the subtasks, $r^{\text{cmpl}}(t_n)$. These two factors form  $\bm{R}^{\text{cmp}}(t_n) = \{r^{\text{CPU}}(t_n), r^{\text{cmpl}}(t_n)\}$. For $\bm{R}^{\text{efc}}$, it is set to 1 if all subtasks executed within time slot $t_n$ yield correct results; otherwise, it is set to 0 if any subtask produces an incorrect result. After executing 2000 tasks, the collected data is merged, and the top 4000 most similar entries are selected using the $K$-means clustering algorithm. These entries are randomly paired to form $Q$ data pairs, labelled with ground truth pairing information $y_q = 1$ to indicate similarity. All data pairs are then transformed into corresponding ACFG pairs.

To simulate anomalies in communication and computing resources, third-party applications are activated to increase CPU usage above $85\%$, while data packets are randomly dropped on the collaborator side. Task completion anomalies are simulated by modifying the task results. The total number of generated anomalous data is $S=1000$. Each anomalous data entry is then randomly paired with one from $Q$, generating an anomalous data pair with ground truth pairing information $y_q = -1$, indicating dissimilarity. All anomalous data pairs are then transformed into their corresponding ACFG pairs. The proposed SRCTE is trained by the mixture of dataset $Q$ and dataset $S$.

\begin{figure}[!]
\centering
\includegraphics[scale=0.55]{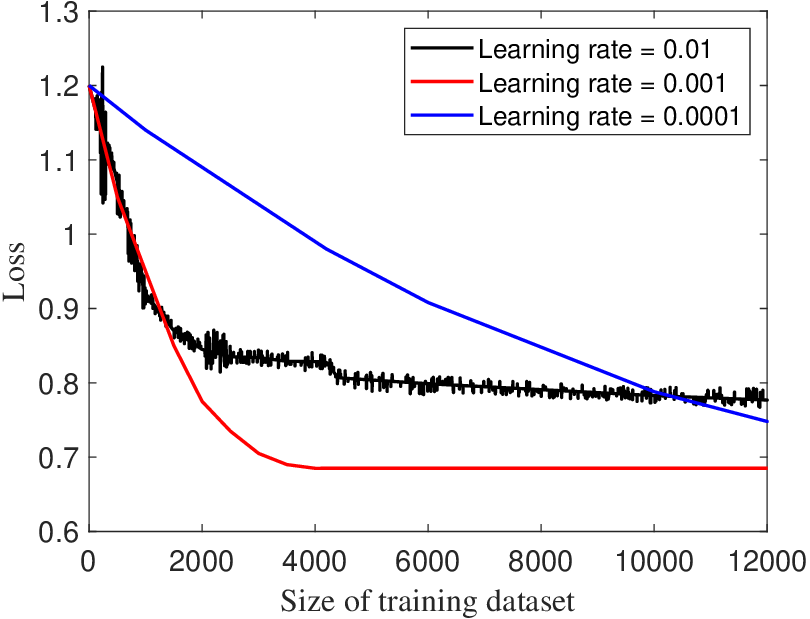}
\caption{Training loss of the proposed SRCTE at different learning rates.}
\label{loss}
\end{figure}

\subsection{Convergence evaluation of SRCTE}

In this subsection, the convergence of the proposed SRCTE is assessed. As illustrated in Fig.~\ref{loss}, when the learning rate is 0.001, SRCTE quickly converges to the global optimum at around 4000 data points. This suggests that effective training can be achieved with a relatively small dataset, which is particularly advantageous in scenarios with limited data collection, enabling quicker deployment in practical applications without compromising performance. Furthermore, this rapid convergence suggests that the model is capable of effectively learning the underlying patterns within the data, ensuring reliable outcomes even with fewer training samples.

\subsection{Optimal selection of embedding layer $H$}

\begin{figure}[!]
\centering
\includegraphics[scale=0.588]{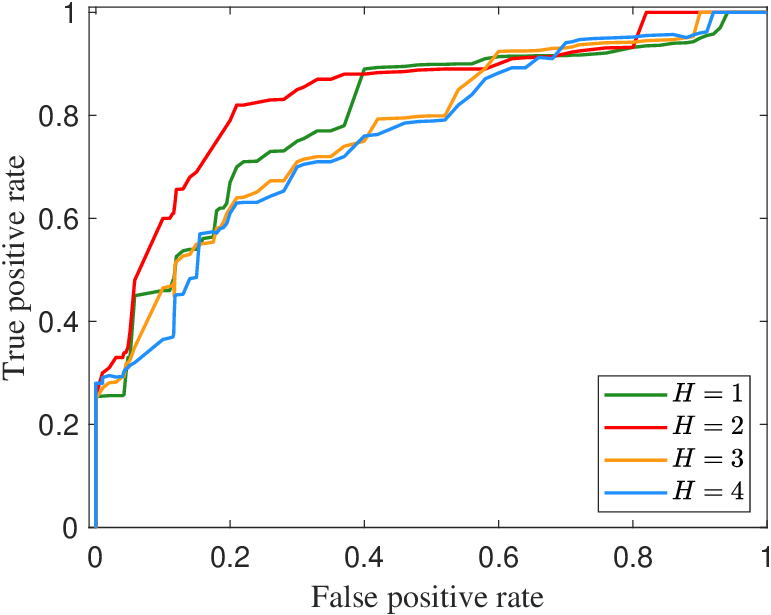}
\caption{Comparison of the impact of the embedding layer $H$ on SRCTE performance.}
\label{roc}
\end{figure}

In this subsection, the impact of embedding layer $H$ on the final results is investigated. A new test dataset containing 1000 data entries is generated based on the method described in Section~\ref{datasetcollection}. The standard receiver operating characteristic (ROC) curve is used to evaluate the performance. The area enclosed by the ROC curve and the $x$-axis is known as the area under the curve, representing the probability that a randomly selected positive example is ranked higher than a randomly selected negative example. As shown in Fig.~\ref{roc}, it is evident that the curve for $H=2$ lies above the curves corresponding to the other three values. This means the proposed SRCTE can achieve a higher true positive rate with a lower false positive rate when $H=2$. This further indicates that when $H=2$, the generated embeddings capture the semantic information of ACFG more effectively. However, when $H$ exceeds 2, it does not improve SRCTE's accuracy and also increases the computational overhead.

\subsection{Continuous trust evaluation comparison}

\begin{figure}[!]
\centering
\includegraphics[scale=0.588]{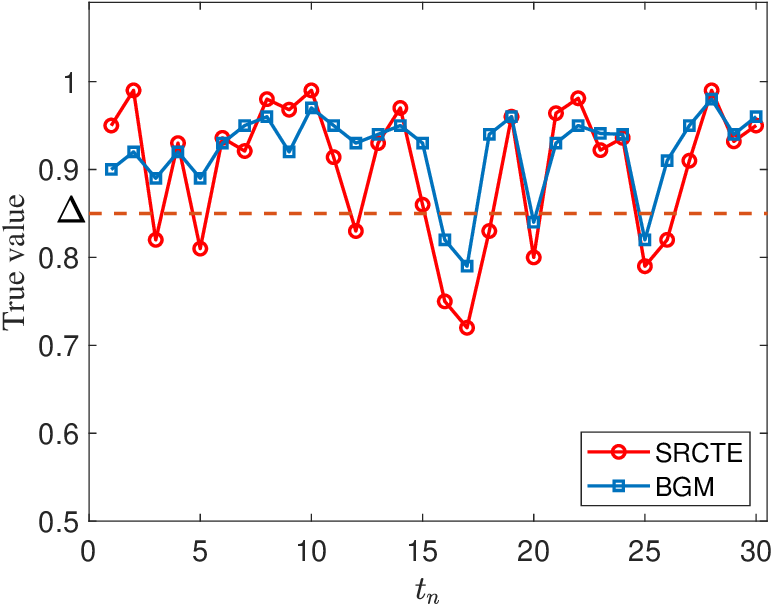}
\caption{Comparison of trust anomaly detection results through continuous trust evaluation.}
\label{comparison}
\end{figure}

In this subsection, we compare the results of continuous trust evaluation with the bipartite graph matching (BGM)~\cite{b20}. The test dataset contains 30 time slots, 10 of which are trust anomalies. The threshold $\Delta$ is set to 0.85. If the trust value of a time slot is below $\Delta$, the collaborator is considered untrustworthy in that time slot. As shown in Fig.~\ref{comparison}, SRCTE successfully detects trust anomalies in 9 time slots, outperforming the comparison algorithm, which identifies only 4. For the one undetected anomaly, the calculated trust value by SRCTE is very close to the threshold. This demonstrates that our proposed SRCTE has a significant advantage in detecting trust anomalies.


\vspace{0.007 in}
\section{Conclusion}
This paper investigated the challenge of rapid and continuous trust evaluation of collaborators to ensure effective task collaboration. To address this issue, a new SRCTE framework was proposed. Initially, the communication and computing resource attributes of the collaborator in a trusted state, along with historical collaboration data, were collected and represented using an ACFG to capture their trust-related semantic information. This ACFG was used as a reference for comparison. During task execution, the collaborator's communication and computing resource attributes, as well as task completion effectiveness, were collected in real-time for each time slot and represented by an ACFG. To capture the deep semantics conveyed by each pair of ACFGs, a Siamese model comprising two shared-parameter Structure2vec networks was employed to generate their embeddings. The similarity calculation was then performed on these embeddings to derive the real-time trust value of the collaborator. A practical system was developed utilizing two Dell EMC 5200 servers and a Google Pixel 8 to evaluate the effectiveness of the proposed framework. Experimental results indicated that SRCTE converges quickly with minimal data and achieves a high anomaly trust detection rate when compared to the baseline algorithm.



\end{document}